# Development of CNN Architectures using Transfer Learning Methods for Medical Image Classification

**Ganga Prasad BASYAL**
College of Business and Information Systems, Dakota State University
Madison, S.D. 57042 USA.

**David ZENG**
College of Business and Information Systems, Dakota State University
Madison, S.D. 57042 USA.

**Bhaskar P. RIMAL**
Become College of Computer and Cyber Science, Dakota State University
Madison, S.D. 57042 USA.

**ABSTRACT**

The application of deep learning-based architecture has seen a tremendous rise in recent years. For example, medical image classification using deep Learning achieved breakthrough results. Convolutional Neural Networks (CNNs) are implemented predominantly in medical image classification and segmentation. On the other hand, transfer learning has emerged as a prominent supporting tool for enhancing the efficiency and accuracy of deep learning models. This paper investigates the development of CNN architectures using transfer learning techniques in the field of medical image classification using a timeline mapping model for key image classification challenges. Our findings help make an informed decision while selecting the optimum and state of the art CNN architectures.

## 1. INTRODUCTION

Computer-aided diagnosis and detection (CAD) have been predominantly applied and researched in the biomedical and medical informatics domain for a long time. Historically the CAD systems were used by pathologists and radiologists in the diagnosis of various diseases through the analysis of medical images [1]. Having huge dependence over the CAD system for a longer period of time allowed CAD systems to set milestones in the field of medical imaging. The researchers have a mixed review on the technology front. Some suggest the methods are efficient enough, whereas few questioned the accuracy of it [1]. Deep learning is another field that successfully emerged in medical image analysis and is widely implemented through its methods and techniques [2]. Deep learning has seen a tremendous increase in its adaptation and application into the medical informatics domain. Having a widespread presence and showing its potential in the last two decades, deep learning has been highly researched and widely applied in the industry. Medical data generated in the form of clinical reports, patient charts, and diagnosis reports result in an exponential surge in medical data. Deep learning can process big data and analyze them efficiently which proved to be an advantage for its rapid and iterative implementation. Deep learning has three key methodologies, namely, Artificial Neural Network (ANN), Recurrent Neural Network (RNN), and Convolutional Neural Network (CNN). ANN and RNN have been applied majorly in text and number data whereas, CNN is applied for vision or image data [3].

CNN gained its attention right from its inception by winning one of the most prestigious challenges for image classification ILSVRC in 2012. It managed to win the competition by a huge margin as compared to the other competitors [2]. Later entering into the biggest medical image classification challenge Multimodal Brain Tumor Segmentation Challenge (BraTS) laid the foundation of CNN into the machine vision domain. Afterward, many other CNN models came into the field (many of which sets the industry standards for others) over time. Medical imaging is a very intriguing field where models are trained to read and analyze images.

Medical image analysis is one of the many application areas of CNN where it is applied for various tasks such as classification, segmentation, detection, and analysis [4]. There are many types of medical images known as image modalities. These include Magnetic Imaging Resonance (MRI) [4], Computed Tomography (CT) [5], X-Ray [6]. Transfer learning on the other had can be defined as the transfer of knowledge gained by one architecture transferred to another domain architecture [7]. Transfer learning provides the advantage of acquiring the domain knowledge for achieving the greater generalization of the model, as the training of the model is done with different sets of data this enables the model to better generalize the data being tested with a different domain.

Furthermore, it seems very difficult for a researcher to decide which CNN architecture to adopt when it comes to medical image classification and the task becomes even more tedious when we have several imaging modalities.

The Permutation and combination of CNN architecture along with image modalities left a huge gap for selecting the optimum choice between these models. The objective of our research is to provide the update and advances in the field of CNN in the last eight years, more specifically its architecture and the application using transfer learning techniques. Our emphasis and focus are on the CNN architecture and its performance over the period of time; therefore, we are paying a bit less attention to the type of medical image modality. In the next section, we perform a timeline mapping of CNN architectural development with respect to the major medical and non-image challenges in the last 8 years.

## 2. DEVELOPMENT OF CONVOLUTIONAL NEURAL NETWORKS

Application of various (CNN) architecture is specific to the task. For example, the study in [6] is a retrospective study where the author compared the results of the CNN model with and without transfer learning and then compared it with a radiologist. The author argues that the results generated by the implementation of VGG-16 CNN architecture provided excellent results and match the human-level performance. Another study in [8] describes CNN architectures using transfer learning with ensemble classifier, the author claims that ensemble is a very efficient method in multi-categorical classification and detection on color fundus images, where images have classified on 16 different species level. A similar method adopted in research in [5] instead of ensemble author integrated two CNN architecture making a pair together, namely, Inception-ResNet and AlexNet-GoogleNet, ResNet with transfer learning performed best among the rest. ResNet and Inception V3 were applied in [9] for bird species classification with the implementation of a two-stage training model achieved an accuracy of 55.67 %, which was one of the best results during the bird species challenge of 2018.

Transfer learning is applied in various CNN architectures, and almost every case, it enhanced the results as compared to the model trained without transfer learning [9], [10], and [11]. Studies in [6] and [12] emphasize that transfer learning increased the accuracy of model performance even having the issue of low sample size. The U-Net architecture applied in [13] and [12] for the segmentation of brain CT images performed very well over the 3D images. The U-Net is efficient in both 2D and 3D medical image segmentation.

## 3. MAPPING THE DEVELOPMENT OF CNN ARCHITECTURE THROUGH IMAGENET CHALLENGE

ImageNet Large Scale Visual Recognition Challenge (ILSVRC) is considered one of the top-ranked image classification challenge. The Challenge consist of 10 million images for training with over ten thousand annotated image (labeled data) categories for the classification and validation and the test dataset consists of over 150 thousand images without labeling [14]. The objective of the challenge is to classify the unannotated natural images into one of the categories of image labels provided at the time of the training dataset. There is three key classification task is given at the challenge: 1) Classification – The output of the model should provide top 5 object categories based on the confidence of the classification, 2) Classification with Localization – Along with classification as in task one the model should also produce five class labels for each image has the bounding box for each image the result should be matched with the actual ground truth, and 3) Fine-grained Classification – This is even a step further in classification, which means subcategory classification i.e., classifying the breed of the dog inside the overall dog category.

In the year 2012, research by Krizhevsky et al., [15] reserved the top spot by introducing the groundbreaking CNN model architecture in the image recognition field, named after the author's first name, i.e., AlexNet. The architecture consists of five convolutional layer and three fully connected layers, whereby the input size of the images is 224x224x3, which then divided into two different layers each representing one GPU (graphics processing unit). The processing of these layers is independent of each other throughout the convolution process and merge in the final dense layer. To overcome the issue of overfitting author utilized two methods 1) Data Augmentation and 2) Dropout with (0.5) rate. The author advocated due to the inception of GPU the model was able to hold the memory for 60 million parameter and 650 thousand neurons, and with this speed, the model was able to achieve the top rank with the lowest error rate of 16% winning by a huge margin over the second-place holder with 26.2% error rate. In 2013, a team of researchers from New York University proposed ZFNet inspired by the first name of the author. It was an alteration of the CNN proposed by [15]. The performance of the proposed model was slightly better than the last year winner and winning over by 4% less error rate with 12%. Another notable CNN architecture proposed by [16] deconvoluted the layer resulting in each layer decomposing the image hierarchically, and reconstructing it again with sparsity constraints.

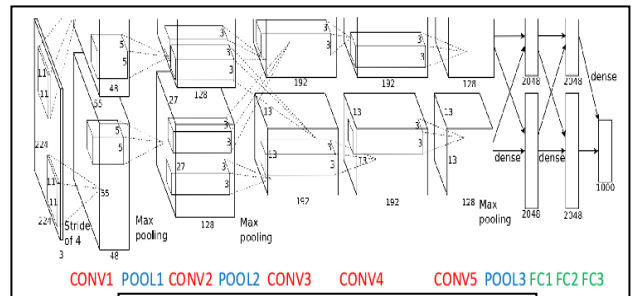

Fig.1 AlexNet Principal Working [15]

A new CNN architecture model called Inception or GoogleNet was introduced [17] in 2014, which has 22 convolutional layers. This model was considered to be the deepest and efficient model at that time. Considering the astounding depth of the model one could argue about the computational efficiency of the model, but researchers had carefully circumvented the issue by inducting 1x1 convolutional for dimensionality reduction, which is stacked on the bottom of traditional convolutional layers, which in terms helped in reducing the computational complexities. Another deep model that gained appreciation and recognition was VGG 19 [18], which is an updated and deeper model than its previous version, i.e., VGG16. The model was able to spot the second-best performance and was very close to the GoogleNet. The author claimed increasing the depth of the architecture had increased the performance as well for the demonstration author compared the architecture with a current and previous winner of the challenge and proved to better in performance in most of the cases. Both the CNN architectures were very efficient with performance benchmarks as low as 6.7 and 6.8 error rates, respectively.

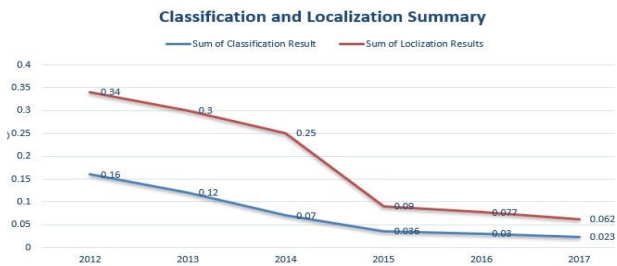

Fig. 2 Performance Measurement of CNN Architecture [19]

Microsoft's research team in the year 2015 coined a novel CNN architecture named ResNet which was based on its principle working of deep residual learning [20]. The architecture was based on the referenced residual learning from the previous layer instead of the unreferenced layers as in other models. The author advocated that the networks created by residual learning are conveniently optimizable and learn even from the deep layers. The winner for 2016 was Gated bi-directional Network (GBD-Net), which achieved the highest mAP rate of 66% [21]. The model was capable of sending the message between the feature in both the processes, i.e., during the feature learning and feature extraction. In 2017, a novel approach had been adopted for video detection and segmentation based on Fast RCNN [22]. This architecture had won the challenge that year. The winner for the classification task was the Squeeze-and-Excitation Network (SE) model [23]. The model utilizes the interdependency feature of the convolutional network through the feature recalibration, by this, the model can learn the high-level feature, and ignore the less information. This was the last year of the challenge as we don't know the specific reason for discontinuing the challenge, but still, we believe that the prime moto could be restructured the data and increase the quality of it.

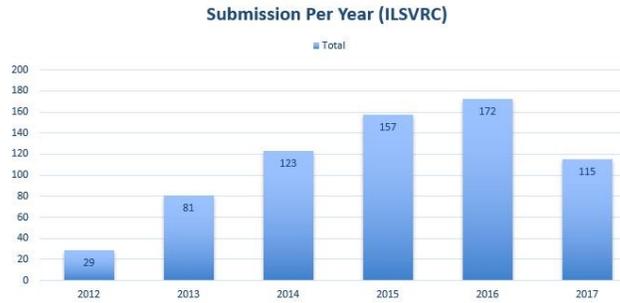

Fig. 3 Number of Submission per year in ILSVRC

**4. MAPPING THE DEVELOPMENT OF CNN ARCHITECTURE THROUGH BRATS IMAGE CLASSIFICATION**

The brain tumor image segmentation (BraTs) challenge is a renowned name in the field of medical image classification used Magnetic Resonance Imagining (MRI) scan, which is publicly available for high-quality research [24]. Starting from the year 2012 in association with the medical image computing and computer-assisted intervention society (MICCAI) conference where a researcher from all around the world has been given a real-world medical image dataset with a problem statement of brain MRI segmentation. Classification or segmentation of brain tumor tissues broadly classified into malignant considered as high-grade tumor and benign tumor considered as low-grade tumor [25]. That can be grouped into three key categories of segmentation: a) manual segmentation, b) semi-automatic segmentation, and c) automatic segmentation methods [26]. CNN architecture excels in the latter two methods as neither it requires large training data nor completely manually annotated data.

In the year 2012, the challenge took place the first time and had received several submissions ranging from researchers from prestigious universities to medical hospitals and healthcare organizations. There was no live conference submission using CNN methods at the time due to the focus of the challenge on multi-atlas labeling rather than the actual segmentation as a whole [27]. The techniques such as Random Forest (RF) and Logistic regression were used those requiring the manual annotation. But after the challenge attempts had been made to analyze the dataset using CNN methods on the 2013 and 2014 challenge [28],[29], where the very initial CNN model was used. The initial models used were shallow such as research by [28] implemented 5 layers in which the first and last layers were the input and output layers, and the middle three-layer included two filter layers and one max-pool layer. The later version of CNN was much advanced, and deeper having many hidden layers.Due to the excellent performance of CNN-based architecture in 2014, the adoption of CNN based model increased in 2015 and major participates have used the CNN model along with traditional manual annotation methods such as RF and logistic regression (LR). The top-ranked dice score was achieved by Havaei et. al [30]. The author applied Input Cascade CNN based on 2D CNN model architecture with entails two pathways covering the small details of the images along with the larger context. The performance of CNN architecture was better than the winner of last year.

With 18 submissions for the BraTs 2016 competition, the bar had been rising and the number of participations made during the year had also reached more than 150% compared to previous years' submission. Though having a higher participation rate still there are participations having traditional methods such as RF and support vector machines (SVM) [31]. With more than half of the submission having some CNN, the application makes it the best choice for the researcher and participants. CNN architecture introduced by Kamnitsas et al. [32] had performed exceptionally well and this is the first time when a 3D CNN architecture had been introduced in competition and won it. Another notable contribution was from Randhawa et al in which they presented a deeper version of 2D CNN having the advantage of the choice over the loss function which uplifted the model performance from earlier versions [33].

The year 2017 had seen exponential growth in the number of participants researching up to over 50 submissions [34]. The winner of the competition, Kamistsas et al. [35], introduced an ensemble model EMMA, comprises of three separates model (U-Net, DeepMedic and Fully Convolutional Neural Network (FCN), using transfer learning method and trained on completely different data the model achieved best results through averaging the output of all three models to get a generalized output. EMMA performed very well yielding a dice score of .90 enhanced, .82 Whole, .75 Core. Runner up for the competition was the Cascaded Anisotropic CNN model introduced by [36] achieving the Dice score of 0.764 enhanced, 0.897 whole, and 0.825 tumor core. The CNN models used in the challenge were very deep having layers as deep as 26 [34].

Furthermore, the succession of the challenge attracts more participants in the year 2018 with numbers exceeding 60, participants from all over the world shown their enthusiasm, and 50 plus submission had utilized CNN architecture as their choice for the classification and segmentation, looking at the model adoption U-Net seems to be the unanimous choice for the researcher as more than half of the researcher used U-Net either standalone or in combination with other CNN architectures [37]. The highest Dice scores were achieved by Myronenko et al. [38]. The author introduced a semantic segmentation network where the input image is reconstructed with the help of auto-encoder which results in the regularization of decoder resulting in efficient model requiring less training samples. The Dice scores reported as the model output were 0.823 enhanced tumors, 0.91 whole tumors, and 0.866 tumor core. Inclusion of modified version of CNN architecture such as ResNet [38], [39] and DenseNet [20], [40]. The runner-up for the challenge utilized a modified version of the U-Net model, author added region-based training along with minor tweaks in postprocessing and loss function computation through which they were able to achieve a dice score of 77.88 enhanced, 87.81 whole, and 80.62 tumor core.

Fig.1 below represents the average Dice Score with the Enhanced tumor, wide tumor, core tumor respectively. Looking at the graph we can ascertain that the graph has an upward trend after the year 2013 which means the score has improved every year as compared to the previous year. The difference in performance between the year 2012 and 2013 is due to the newly introduced 2D CNN which got improved year after year resulting in the best choice for medical image classification.

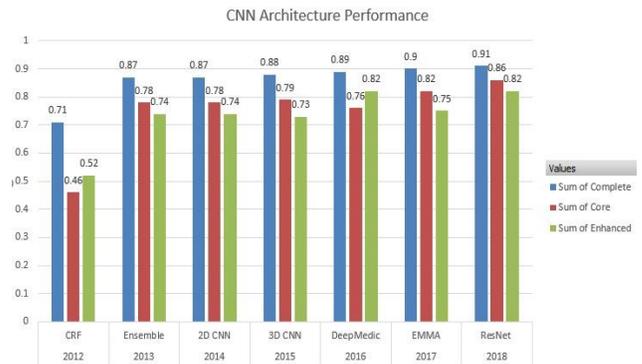

Fig.4 Performance of CNN Architectures in BraTs Challenge

Fig.2 provides information on the number of submissions per year. The year 2017 and 2018 had witnessed the highest number of submissions till then.

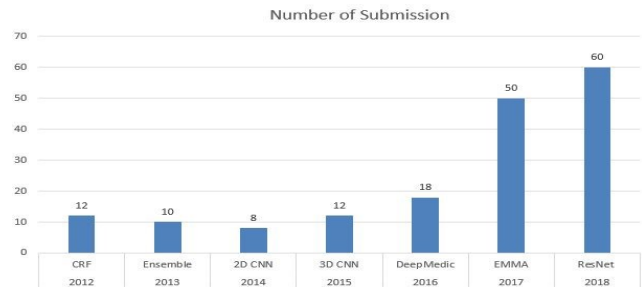

Fig.5 Number of Submission per year

## 5. DISCUSSION AND FUTURE WORK

ImageNet Challenge being there for a longer period had an edge over BraTs for the early introduction of the initial CNN model. Looking at the empirical analysis and our literature review we identified that CNN architecture such as ResNet and U-Net was first introduced to BraTs and later their improved version introduced to ImageNet

Challenge [41],[18]. Furthermore, CNN architecture such as AlexNet being introduced to ImageNet first, and later being applied to medical image challenge [41]. But one pattern is common despite the platform. These CNN models got better and better as their newer, and improved versions had been introduced.

State-of-the-Art improved versions of prior CNN models such as EfficientNet [42], NAS U-Net [43], and NAS V-Net [44] are the latest models. Deploying these models to the prior datasets results in achieving astonishing results, surpassing all of the previous models. EfficientNet is the improved version of ResNet and MobileNet, it is trained on low parameters and yielded very good results [42]. It tends to be one of the best models for natural image classification in the current scenario. The model has been implemented only for natural image classification but has not been applied to medical image classification yet. Natural Architecture Search (NAS) models such as NAS U-Net and V-NAS had achieved promising results in medical image classification surpassing all the previous standards in medical image classification. Based on the prior results generated by different ensemble learning models such EMMA and (CRF+ RF) lead them to win the competitions in past. These ensemble models used contemporary CNN architectures. Our future research work will be focused on deploying the three latest and state-of-the art CNN models i.e. (NAS U-NET, V-NAS and Efficient Net). Implementing them in the form of multi-stages followed by averaging the results in the form of ensemble learning for the medical image classification.

## 6. CONCLUSION

In our review research, we discussed the fundamentals of deep learning models and transfer learning techniques. To the best of our knowledge, this is a unique study attempted to provide a comprehensive review of CNN architectures and of their performance through visualization, encapsulating over a half-decade time frame. We structured them in the form of timeline mapping development using two of the biggest image challenge competitions. We reviewed some of the foundational CNN models, such as AlexNet, VGG, Inception (GoogleNet), GBDNet, and ResNet, and compared their performance with their counterparts. Our study suggests that ensembles of CNN models had performed exceptionally well in both the source and target domains of image recognition. The inclusion of the transfer learning method along with mechanisms such as normalization and augmentation, can substantially increase the training efficiency and the model performance.